\documentclass[lettersize,journal]{IEEEtran}
\usepackage{graphicx}
\newcommand{\fref}[1]{Fig. \ref{#1}}
\newcommand{\sref}[1]{Section \ref{#1}}
\newcommand{\tref}[1]{TABLE \ref{#1}}
\newcommand{\eref}[1]{Eq. (\ref{#1})}
\usepackage{array}
\usepackage{amsmath}
\usepackage{bm}
\usepackage{amssymb}
\usepackage{algorithm}
\usepackage{algorithmic}

\usepackage{array}
\usepackage{multirow}
\usepackage{booktabs}
\usepackage{graphicx}
\ifCLASSOPTIONcompsoc
\usepackage[caption=false, font=normalsize, labelfont=sf, textfont=sf]{subfig}
\else
\usepackage[caption=false, font=footnotesize]{subfig}
\fi
\usepackage{url}
\usepackage{color,xcolor}
\begin{document}

\title{Learning to Branch in Combinatorial Optimization with Graph Pointer Networks}
%
%
%

\author{Rui~Wang, ~\IEEEmembership{Senior Member, IEEE},
	Zhiming~Zhou,
	Tao~Zhang,
	Ling~Wang,
	Xin~Xu,
	Xiangke~Liao,
	
	Kaiwen~Li
	\thanks{This paper is partially supported by the National Science Fund for Outstanding Young Scholars (62122093) and the National Natural Science Foundation of China (No. 72071205).}
	
	\thanks{Rui~Wang, Kaiwen Li (corresponding author), Tao~Zhang are with the College of Systems Engineering, National University of Defense Technology, Changsha 410073, PR China, and also with the Hunan Key Laboratory of Multi-Energy System Intelligent Interconnection Technology, HKL-MSI2T,Changsha 410073, PR China. (e-mail: ruiwangnudt@gmail.com, kaiwenli\_nudt@foxmail.com, zhangtao@nudt.edu.cn); Zhiming Zhou is with Institute of Automation, Chinese Academy of Sciences, Beijing, 100190, P. R. China. (zhiming.zhou@ia.ac.cn); 
		Xin~Xu is with the College of Intelligence Science and Technology, National University of Defense Technology, Changsha 410073, PR China;
		Ling Wang is with Department of Automation, Tsinghua University, Beijing, 100084, P. R. China. (wangling@mail.tsinghua.edu.cn);
		Xiangke~Liao is with the College of Computer Science and Technology, National University of Defense Technology, Changsha 410073, PR China.
	}

\thanks{Manuscript received March 19, 2022; revised March 26, 2021.}}

%
%

\markboth{Journal of IEEE/CAA Journal of Automatica Sinica, Vol. 00, No. 0, Month 2023}
{Rui Wang \MakeLowercase{\textit{et al.}}: Learning to Branch in Combinatorial Optimization with Graph Pointer Networks}

\maketitle

\begin{abstract}
Branch-and-bound is a typical way to solve combinatorial optimization problems. This paper proposes a graph pointer network model for learning the variable selection policy in the branch-and-bound. We extract the graph features, global features and historical features to represent the solver state. The proposed model, which combines the graph neural network and the pointer mechanism, can effectively map from the solver state to the branching variable decisions. The model is trained to imitate the classic strong branching expert rule by a designed top-k Kullback-Leibler divergence loss function. Experiments on a series of benchmark problems demonstrate that the proposed approach significantly outperforms the widely used expert-designed branching rules. Our approach also outperforms the state-of-the-art machine-learning-based branch-and-bound methods in terms of solving speed and search tree size on all the test instances. In addition, the model can generalize to unseen instances and scale to larger instances.

\end{abstract}


\IEEEpeerreviewmaketitle

\section{Introduction}

\IEEEPARstart{C}{ombinatorial} optimization seeks to explore discrete decision spaces, and finds the optimal solution in acceptable execution time. Combinatorial optimization problems arise in diverse real-world domains such as manufacturing, telecommunications, transportation and various types of planning problem \cite{festa2014brief, schrijver2005history}. This kind of problems can be immensely difficult to be solved, since it is computationally impractical to find the best combination of the discrete variables through exhaustive enumeration. Actually most of the NP-hard problems in mathematical and operational research fields are typical examples of combinatorial optimization, such as Traveling Salesman Problem (TSP), Maximum Independent Set \cite{abe2019solving}, Graph Coloring \cite{yolcu2019learning}, Boolean Satisfiability \cite{yolcu2019learning}, etc..

A vast of approaches have been proposed to tackle combinatorial optimization challenges these years. They can be basically divided into the following categories: exact algorithms, approximation algorithms and heuristics. Exact algorithms are algorithms that can always find the optimal solution to a combinatorial optimization problem. A naive way is searching all possible solutions through enumeration, however, costing intractable solving time. Some advanced techniques have been proposed, such as branch-and-bound, to efficiently prune the searching space. Approximation algorithms, in some cases, can solve an optimization problem in polynomial-time, and can provide a theoretically guaranteed bound on the ratio between the obtained solution and the optimal one. However, such algorithms may not exist for all real-world combinatorial optimization problems. Heuristics provide no guarantees for the solution quality, but are faster than the above approaches. Hard-won expertise and trial-and-error efforts are often required to design the heuristics.

As exact algorithms can always solve an problem to optimality, and no problem-specific heuristic requires to be hand-crafted, modern optimization solvers generally employ exact algorithms, typically the branch-and-bound ($\mathrm{B}\&\mathrm{B}$) approach, to solve the combinatorial optimization problems, which can be formulated as mixed-integer linear programs (MILPs). $\mathrm{B}\&\mathrm{B}$ solves general MILPs in a divide-and-conquer manner. $\mathrm{B}\&\mathrm{B}$ \cite{land2010automatic} recursively splits the search space of the problem into smaller regions in a tree structure, where each node represents the subproblem that searches subsets of the solution set. Subtrees can be pruned once it provably cannot produce better solutions than the current best solution; otherwise, the subtree is further partitioned into subproblems until an integral solution is found or the subproblem is infeasible. In this solving process, there are several decision-making problems that should be considered to improve the performance: \emph{node selection problem}, i.e., which node/subproblem should we select to process next given a set of leaf nodes in the search tree?; \emph{variable selection problem} (a.k.a. branching), i.e., which variable should we branch on to partition the current node/subproblem?

For a long time, such decisions are made according to some carefully designated heuristics on a specific type of MILP instances. A lot of designing and trial-and-error efforts are required to hand-craft these hard-coded expert heuristics. In recent years with the development of artificial intelligence, more attentions are drawn on learning the heuristics by machine learning models instead of designing by experts. This idea makes sense since heuristics are typically formed by a set of rules, which can be possibly parameterized by models, such as the deep neural networks. Such learning-based approaches have been investigated in recent years \cite{He2014, Khalil2016, gasse2019exact,Gupta2020}. However, this line of work still raises the following challenges: how to extract effective features to represent the current state of the $\mathrm{B}\&\mathrm{B}$ process, based on which the branching decision is made; how to design effective models to map from the $\mathrm{B}\&\mathrm{B}$ state to the branching decision.

In this paper we propose a graph pointer network model to address the above challenges. In specific, we focus on the branching problem, i.e., which variable to branch on. Instead of designing the branching heuristics manually for each problem type, we propose to learn the branching heuristics automatically by a novel model to reduce the solving time of MILPs. We achieve this by using imitation learning to approximate the \emph{strong branching} branching rule, which is empirically effective but computationally expensive. Though this idea is not new \cite{Khalil2016, gasse2019exact,Gupta2020}, we improve the performance of the learning model in a novel way. The contributions are as follows:

\begin{itemize}
\item In addition to graph features, we design the global and historical features to represent the solver state. The extracted features can provide a richer representation for the problem state.
\item We develop a new model that combines the graph neural network and the pointer mechanism. Graph neural network is used to encode the graph features, and the pointer mechanism is used to incorporate the global and historical features to output the variable index.
\item A top-k Kullback-Leibler divergence loss function is designed to train the model to imitate the expert branching rules.
\item The proposed approach can outperform expert-designed branching rules and state-of-the-art machine learning methods on all the test problems.
\item Once trained, the model can generalize to unseen larger instances.
\end{itemize}

\section{Related work}

Recent days have seen a surge of applying artificial intelligence methods for combinatorial optimization.

Vinyals et al. \cite{vinyals2015pointer} developed a pointer network model for solving small scale combinatorial optimization problems like the traveling salesman problems (TSPs). It borrowed the idea of the widely used sequence-to-sequence model in the machine translation field, and used the attention mechanism to map from the input sequence to the output sequence. This work inspired a number of subsequent researches that use machine/deep learning methods for combinatorial optimization.

Most the current works focus on solving the combinatorial optimization problems in an end-to-end manner. Bello et al. \cite{bello2016neural} first proposed to use a deep reinforcement learning (DRL) method to optimize the pointer network model, which can output the solution sequence directly. Nazari et al. \cite{nazari2018deep} investigated the vehicle routing problem (VRP) by modifying the pointer network and the attention mechanism. Khalil et al. \cite{khalil2017learning} developed a \emph{structure2vec} graph neural network (GNN) model for combinatorial optimization. The GNN model can encode the graph feature of the problem and aid the decisions. Other works \cite{mittal2019learning,nowak2017note,joshi2019efficient, li2018combinatorial} explored advanced GNN models like the graph convolution networks (GCNs) and diverse training methods to solve the combinatorial optimization problems more effectively. Moreover, authors in \cite{deudon2018learning, kool2018attention} improved the attention mechanism of the pointer network by leveraging the recent advances of the famous Transformer model \cite{vaswani2017attention} in the field of seqence-to-seqence learning. The attention model developed by Kool et al. \cite{kool2018attention} achieved the state-of-the-art performance among the above approaches. This model can solve a number of combinatorial optimization problems, such as the TSP, VRP, the Orienteering Problem, etc. In addition, Li et al. \cite{li2020deep} extended this line of work to a multiobjective version.

Regarding using the artificial intelligence methods to improve the $\mathrm{B}\&\mathrm{B}$ algorithm for combinatorial optimization, Bengio et al. \cite{Bengio2018} made a thorough survey for this line of works. He et al. \cite{He2014} developed a DAgger model to learn the node selection strategy by imitation learning. On the contrary, Khalil et al. \cite{Khalil2016} focused on the variable selection problem, and developed a machine learning model to mimic the classic strong branching strategy. Extensive features of the candidate branching variables are carefully extracted as the input of the model. The model is trained by minimizing the difference of the predicted branching decisions and the decisions made by the strong branching. Moreover, Gasse et al. \cite{gasse2019exact} developed a novel GCN model for learning the variable selection strategy. They exploited the variable-constraint bipartite graph feature of the mixed-integer linear programs (MIPs), and encoded the branching strategy into a graph neural network. The model is trained to mimic the strong branching policy by imitation learning. Following this work, Gupta et al. \cite{Gupta2020} designed a hybrid model that uses the above GCN model at the root node and a weak but fast model at the the remaining nodes. This method has a weaker predictive performance but an overall faster solving speed due to its less computational cost. In addition, Nair et al. \cite{2020Solving} proposed two models, \emph{Neural Diving} and \emph{Neural Branching}, to enhance the traditional MIP solver. \emph{Neural Diving} predicts the partial assignments for its integer variables, which can result smaller MIPs. \emph{Neural Branching} learns a neural network-based variable selection policy that can reduce the overall solving time.

The remainder of the paper is organized as follows. \sref{pre} introduces the preliminaries of the work. The proposed graph pointer network model is described in \sref{s:model}. \sref{sec:policy-learning} outlines the imitation learning method for optimizing the model parameters. The experiment setup and numerical results are presented in \sref{sec:experiments}. The last section gives some concluding remarks and future perspectives.

\section{Preliminaries}
\label{pre}

\subsection{Problem Definition}

\textbf{Mixed Integer Linear Program}

 A combinatorial optimization problem can be always modeled as a mixed integer linear programming problem (MILP), having the form:

\begin{equation}
\begin{aligned}
\min \quad & \mathbf{c^{\top} x}\\
\textrm{s.t.} \quad & \mathbf{A x} \leq \mathbf{b}\\
  & \mathbf{l} \leq \mathbf{x} \leq \mathbf{u}    \\
  & x_{i} \in \mathbb{Z}, \quad i \in \mathcal{I} \\
  & \mathbf{c} \in \mathbb{R}^{n}, \mathbf{A} \in \mathbb{R}^{m \times n}, \mathbf{b} \in \mathbb{R}^{m}, \mathbf{l, u} \in \mathbb{R}^{n},
\end{aligned}
\label{milp}
\end{equation}

where the aim is to find an optimal set of $\mathbf{x}$ to minimize the objective function with $\mathbf{c}$ as the objective coefficient vector. There are $m$ constraints and $n$ decision variables. A subset of the decision variables are integer, and $\mathcal{I} \subseteq\{1, \ldots, n\}$ is their index set. $\mathbf{A,b}$ are the coefficient matrix and the right-hand-side vector of the constraints. $\mathbf{l, u}$ bound the decision variables.

\textbf{LP relaxation of a MILP}

An MILP can be relaxed to a linear program (LP) by eliminating all the integrality constraints. In the minimization style, the solution obtained by solving the LP relaxation of the \eqref{milp} provides a lower bound to \eqref{milp}.

 \textbf{Branch-and-bound}

Branch-and-bound begins by solving the LP relaxation of the original MILP. The obtained solution $\mathbf{x^*}$ provides the lower bound to the problem. If the obtained solution respects all the MILP integrality constraints, it is the optimal solution to \eqref{milp}, and the algorithm terminates. If not, the LP relaxation is further partitioned into two subproblems by branching on an integer variable that does not respect integrality of the MILP. This is done by adding the following two constraints into the LP relaxation, respectively \cite{gasse2019exact}:
\begin{equation}
x_{i} \leq\left\lfloor x_{i}^{\star}\right\rfloor, x_{i} \geq\left\lceil x_{i}^{\star}\right\rceil, \quad \exists i \in \mathcal{I} \mid x_{i}^{\star} \notin \mathbb{Z},
\end{equation}
where $\left\lfloor x_{i}^{\star}\right\rfloor$ refers to the maximum integer value that is smaller than $x_{i}^{\star}$, and $\left\lceil x_{i}^{\star}\right\rceil$ is the minimum integer value that is larger than $x_{i}^{\star}$. Here $i$ is called the branching variable.

By branching on $i$, two new LPs are constructed, which refer to the leaf nodes/subproblems of the search tree. The next step is to pick one leaf node, and repeat the above steps. Once a \emph{feasible} solution $\hat{x}$ is found, that is, all the MILP integrality constraints are satisfied, it provides the upper bound to the problem. If a solution is found with a lower objective value than $\hat{x}$, the upper bound is updated. On the other hand, if a solution is found with worse objective value than the current upper bound, this subproblem is pruned and no longer branched. The subproblem is also fathomed if the solution is integer or the LP is infeasible. The above procedures repeat until no subproblems remains. The incumbent solution with the best bound is returned \cite{gasse2019exact}.

\subsection {Branching strategies} \label{sec:rules}

In the \emph{branching variable selection} decision process, an integer variable $i$ is selected among the candidate variables $\mathcal{C}=\left\{i \mid x_{i}^{\star} \notin \mathbb{Z}, i \in \mathcal{I} \right\}$ that do not satisfy the integer constraint. Existing methods usually score each candidate variable in $\mathcal{C}$ according to some handcrafted heuristics, and the variable with the largest score is selected for branching. The most commonly used scoring criterion is the change of the lower bound of the sub-problem after the variable is branched. Based on this criterion, a series of branch rules are designed to improve the efficiency of $\mathrm{B}\&\mathrm{B}$.

Strong branching (SB) is an effective but expensive scoring heuristic, which is found empirically that, it can always produce the smallest $\mathrm{B}\&\mathrm{B}$ search tree compared with other heuristics \cite{Khalil2016}. SB rule explicitly measures the upper and lower bounds changes of the sub-problem, so as to select the best branching variable, which is computed as follows. For the LP sub-problem corresponding to the current node $N$, its LP solution is $\mathbf{x^*}$, and its corresponding objective value is $z^*$. By branching on variable $i$, two LP sub-problems $N_i^-$ and $N_i^+$ are obtained, and the corresponding objective values are $z_i^{*-}$ and $z_i^{*+}$. If $N_i^-$ and $N_i^+$ have no feasible solutions, then $z_i^{*-}$ and $z_i^{*+}$ are set to very large values. Therefore, the change of the objective function value after branching on variable $i$ is $\Delta_i^-=z_i^{*-}-z^*$ and $\Delta_i^+=z_i^{*+}-z^*$. The SB score is calculated as \cite{Khalil2016}:
\begin{equation}
	S B_{j}=\operatorname{score}\left(\max \left\{\Delta_{j}^{-}, \epsilon\right\}, \max \left\{\Delta_{j}^{+}, \epsilon\right\}\right)
\end{equation}
where the product function is usually considered as the scoring function, that is, $\operatorname{score}\left(a,b\right)=a\times b$. SB rule computes the SB scores for all the candidate variables in the candidate set $\mathcal{C}$, and selects the decision variable with the largest SB score to branch on. Each branching decision requires long computational time since computing each SB score requires solving two LP sub-problems. In this case, the SB-based $\mathrm{B}\&\mathrm{B}$ algorithm usually suffers heavy computational burden although SB can greatly reduce the search tree size.

In view of the heavy computational burden of the SB method, calculating the pseudocost instead of the SB score is another commonly used method in the current optimization solver. Pseudocost branching (PB) estimates the score of a variable according to its historical scores during the previous search process. Instead of solving the two sub-problems by branching on $i$, the upwards (downwards) score of variable $i$ is the average value of the objective value changes when upwards (downwards) branching on variable $i$ in the previous branching process. This can greatly shorten the calculation time. Denote the upwards and downwards average scores of variable $i$ as $\Psi_{j}^{-}$ and $\Psi_{j}^{+}$, PC is calculated as \cite{Khalil2016}:
\begin{equation}
	P C_{j}=\operatorname{score}\left(\left(x_{i}^*-\left\lfloor x_{i}^* \right\rfloor\right) \Psi_{j}^{-}, \left(\left\lceil x_{i}^* \right\rceil-x_{i}^* \right) \Psi_{j}^{+}\right)
\end{equation}

where $x_{i}^*-\left\lfloor x_{i}^* \right\rfloor$ and $x_{i}^*-\left\lceil x_{i}^* \right\rceil$ represent the decimal part of the variable value. PC method can effectively reduce the computing time of each branching decision. However, the search tree is much larger than that obtained by SB, since there is no sufficient historical data in the early stage of the searching to estimate the variable socres, which results in incorrect branching decisions. In view of the pros and cons of SB and PC, the reliability branching (RB) method applies SB at the beginning of the search till enough historical data is accumulated, and then applies PB in the subsequent process.

It can be seen that there is a contradiction between the branching performance and the time cost by making each branching decision.
In this study, we aim to use the deep learning method to imitate the SB heuristic, which is good performing but expensive, so as to reduce the computational burden.

\section {Model}
\label{s:model}
We design a graph pointer network (GPN) model to mimic the above-mentioned SB strategy. The input of the model is the current state of the solver, and the output is the variable selection decision. We first formulate $\mathrm{B}\&\mathrm{B}$ as a Markov decision process. At each step, the model perceives the current state and selects the variable. The state of the solver changes accordingly. In addition, we define the state of the solver, including the graph structure feature, global feature and historical feature. Finally, a graph pointer neural network model is designed according to the state definition, which can perceive the current state of the solver and make branching decisions.

\subsection{Markov decision process modeling}\label{sec:mdp}

\begin{figure}
	\centering
	\includegraphics[width=1\linewidth]{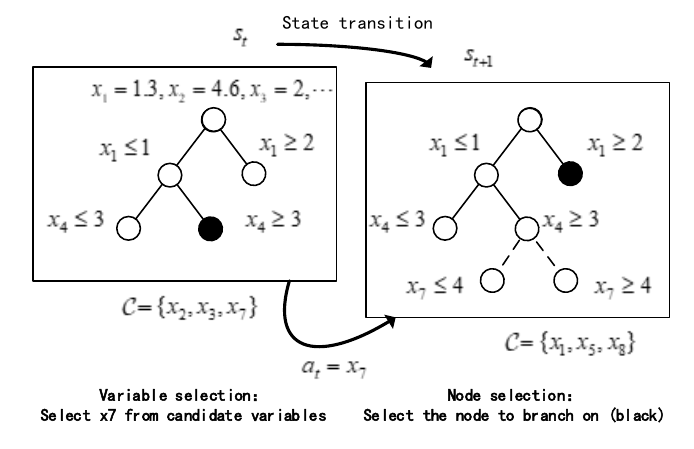}
	\caption{Markov decision process of the branch-and-bound.}
	\label{fig:bb1}
\end{figure}

$\mathrm{B}\&\mathrm{B}$ can be modeled as a Markov decision process \cite{gasse2019exact}, as shown in \fref{fig:bb1}.

At each decision step $t$, the current state of the solver is $\mathbf{s}_t$, which represents the state of the current search tree. Based on the current state of the solver $\mathbf{s}_t$, the agent selects a variable $a_t = i$ from the candidate set $\mathcal{C}=\left\{i \mid x_{i}^{\star} \notin \mathbb{Z}, i \in \mathcal{I} \right\}$ according to the strategy $\pi(\mathbf{a}_t\,|\,\mathbf{s}_t)$.

The solver solves the two LP sub-problems after branching on variable $i$. Subsequently, the solver updates the upper and lower bounds, prunes the search tree, and selects the next leaf node to branch. At this time, the solver has been converted to a new state $\mathbf{s}_{t+1}$. Then the solver applies the branch strategy $\pi(\mathbf{a}_{t+1}\,|\,\mathbf{s}_{t+1})$ again to make the branching decision. This process is looped until all the leaf nodes are explored.

The initial state of the Markov decision process corresponds to the root node of the $\mathrm{B}\&\mathrm{B}$ search tree. And the final state is the end of the optimization process, i.e., all leaf nodes cannot be branched further. Denote the branching strategy as $\pi$, the Markov decision process can be modeled as \cite{gasse2019exact}:
\begin{equation*}
	p_\pi(\tau) = p(\mathbf{s}_{0}) \prod_{t=0}^{T-1}\sum_{\mathbf{a} \in \mathcal{A}(\mathbf{s}_t)} \pi(\mathbf{a}\,|\,\mathbf{s}_{t}) p(\mathbf{s}_{t+1}\,|\,\mathbf{s}_{t}, \mathbf{a})
	\text{.}
\end{equation*}

In this paper, we learn the branching strategy $\pi$ to imitate the SB rule, which is realized through the following steps: 1) Define the problem state $s_t$. At each step of the branch decision, branch decision needs to be made according to the current problem state. However, there is no standardized definition of the solver state. It is necessary to extract effective features to better represent the solver state, so as to make better decisions accordingly. 2) Parameterize the branch strategy $\pi$ via a novel model. The model should be able to map the problem state $s_t$ to the branching action $a_t$ correctly. The models, such as neural networks, random forests and support vector machines, need to be designed according to the characteristics of $\mathrm{B}\&\mathrm{B}$. 3) Optimize the parameters of the model by an effective training algorithm. The model $\pi$ can be learned through a variety of machine learning methods to minimize the size of the search tree or reduce the total run-time of the $\mathrm{B}\&\mathrm{B}$ algorithm.

The proposed deep learning-based $\mathrm{B}\&\mathrm{B}$ method is constituted of the above three parts introduced as follows.

\subsection {State Definition}
\label{sec:state}
We first define the state $s_t$ of $\mathrm{B}\&\mathrm{B}$ at the decision-making step $t$.  In addition to the graph features introduced in \cite{gasse2019exact}, we further design the global features and historical features of the problem, which can provide a more thorough representation of the solver state. Therefore, $s_t$ is composed of variable features, constraint features, edge features, global features, and historical features, namely $s_t = \left( V,C,E,G,H\right)$.

\begin{figure}
	\centering
	\includegraphics[width=0.7\linewidth]{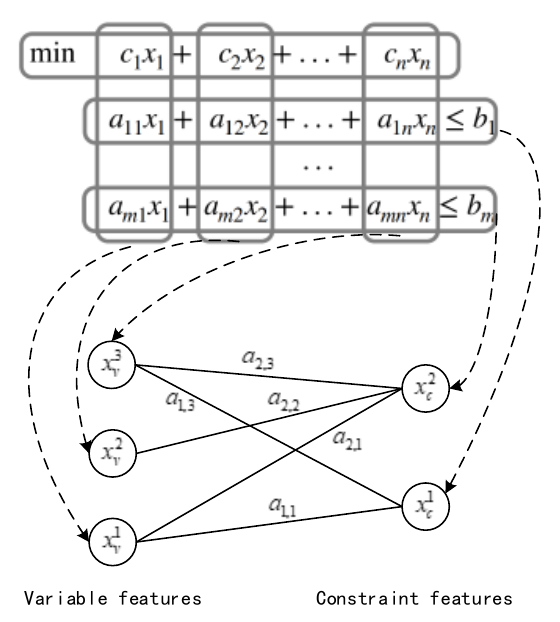}
	\caption{Graph structure of the MILP state.}
	\label{fig:bb2}
\end{figure}

The graph features $\left( V,C,E\right)$ of the problem is defined by the bipartite graph of the current solver state, as shown in \fref{fig:bb2}. The bipartite graph is composed of $m$ constraints and $n$ variables. Variables $x_1, x_2, \cdots, x_n$ are on the left side of the graph. The right-hand side (constant) term of the constraint is on the right side of the graph. The edge $(i,j) \in E$ of the graph is the connection of the variable $i$ and the constraint $j$, i.e., whether the constraint $j$ includes the variable $i$. The weight of the edge is the coefficient of the variable $i$ in constraint $j$.

According to the bipartite graph structure, we define the solver state which is composed of variable features, constraint features, edge features, global features, and historical features:
(1) Variable features represent the attributes of candidate variables at branching step $t$, including the variable type, variable coefficient, current value of the variable, whether the current value of the variable is on the boundary, the decimal part of the solution value of the variable, etc. There are $n$ candidate variables in total, and the feature dimension is $d$. Therefore, the variable feature dimension is $n \times d$. The detailed introduction of the variable features is listed \tref{tab:var}.

(2) Constraint features represent the attributes of the LP constraint at branching step $t$, such as the right value of the constraint, whether the left value of the constraint exactly reaches the boundary, the similarity of the constraint coefficient and the target coefficient, etc. The current LP problem has a total of $m$ constraints, and the feature dimension is $c$. Thus, the dimension of the constraint feature is $m \times c$, and the detailed description of the constraint features can be found in \tref{tab:cons}.

(3) Edge feature is the coefficient of each variable in each constraint. Therefore, there are $m \times n$ edges in total, and the feature dimension is 1. The coefficient value is 0 if the constraint does not contain a certain variable.

(4) Global feature $G$ represents the global state of the solver, such as the current optimality gap of the problem, the gap between the objective value of the current node and the upper/lower bounds, the depth of the current search tree, the depth of the current node, etc. We design and extract the global features using the API interface of PySCIPOpt, which is an open source $\mathrm{B}\&\mathrm{B}$ solver. We list the detailed global features in \tref{tab:global}.

$G$ mainly includes two parts: 1) global features of the whole MILP, including the gap between the upper and lower bounds of the current stage of MILP, the number of feasible solutions/infeasible solutions, etc.; 2) global features of the current LP sub-problem node, including the depth of the current node, the LP objective value information of the current node, etc.

The depth of the current node and the gap between the upper and lower bounds can be directly obtained by calling the PySCIPOpt interface. The number of feasible/infeasible solutions is computed by the proportion of leaf nodes that produce feasible/infeasible solutions:
\begin{equation}
	P_{feasible} = \frac{N_{feasible}}{\max (N_{leaves},0.1)}
\end{equation}

The gap between the current node's LP objective value and the global upper/lower bounds $gap$ is calculated by the following formula according to \cite{zarpellon2021parameterizing}:
\begin{equation}
	\operatorname{gap}(x, y)= \begin{cases}0 & , \text { if } x y<0 \\ \frac{|x-y|}{\max \left\{|x|,|y|, 1 \times 10^{-10}\right\}} & , \text { else }\end{cases}
\end{equation}
where the current node's LP objective value and the global upper/bounds are obtained from the PySCIPOpt interface.

The relative position $pos$ of the current node's LP objective value to the global upper/lower bounds is computed by \cite{zarpellon2021parameterizing}:
\begin{equation}
	\texttt{relPos}(z,x,y) = \frac{|x-z|}{|x-y|}.
\end{equation}

(5) Historical feature consists of two parts. The first part is comprised of features of all past branching decisions $\mathcal{C}_1 = a_1 \cdots a_{t-1}$ at previous steps $1 \cdots t-1$. The second part is comprised of features of variables $\mathcal{C}_2 = x_1, x_2, \cdots $ whose values have changed when generating the current node. That is,  $\mathcal{C}_2$ is the set of variables whose values have changed in the solution of the new problem after adding an integer constraint to the parent problem.

Traditional approaches only considers variable features, constraint features and edge features \cite{gasse2019exact}. This work further extracts global features and historical features, so as to obtain a richer representation of the environment state $s_t$. The global status of the current search tree and the current node can provide more information for the agent to make the branching decisions. Moreover, observing the variables whose values have changed when generating the current node, and observing the variables selected during the historical branching process, can also provide effective information for making the branching decisions. Therefore, it is expected that adding additional global features and historical features can better describe the state of the current problem.
\begin{table}[htbp]
	\centering
	\caption{Variable features}
	\begin{tabular}{p{1cm}p{7cm}}
		\toprule
		type  & description \\
		\midrule
		categorical   & variable type, 0: 0-1 binary, 1: integer and 2: continuous \\
		real    & normalized variable coefficient in the objective function \\
		binary   & if the variable owns a upper/lower bound \\
		binary   & if the current solution value of the variable is its upper/lower bound\\
		real    & fractional part of the variable's current solution value\\
		categorical   & 0: the variable is at its lower bound, 1: variable's value lies between the upper and lower bounds (basic), 2: the variable is at its upper bound, 3: rare case \\
		real    & reduced cost, the variable's solution value can become positive if we reduce the objective coefficient of the variable by this value \\
		real    & number of LP iterations since the last time the variable was basic \\
		real    & solution value of the variable at the current node \\
		real    & the variable's value of the best primal solution \\
		real    & average value of the variable in all the feasible solutions found so far \\
		\bottomrule
	\end{tabular}
	\label{tab:var}
\end{table}%

\begin{table}[htbp]
	\centering
	\caption{Constraint features}
	\begin{tabular}{p{1cm}p{7cm}}
		\toprule
		type  & description \\
		\midrule
		real    & similarity between the left-hand-side coefficients of the constraint and the objective coefficients \\
		real    & normalized right-hand-side (constant) value of the constraint\\
		real    & number of iterations since the last time the constraint was active \\
		binary   & dual variable's value of the constraint \\
		binary   & if the constraint is at the bounds \\
		\bottomrule
	\end{tabular}
	\label{tab:cons}
\end{table}%

\begin{table}[htbp]
	\centering
	\caption{Global features}
	\begin{tabular}{p{1cm}p{7cm}}
		\toprule
		type  & description \\
		\midrule
		real    & depth of the current node \\
		real    & normalized number of feasible solutions \\
		real    & normalized number of infeasible solutions  \\
		real    & gap between the global upper and lower bounds \\
		real    & gap between the current node's LP objective value and the global upper bound \\
		real    & gap between the current node's LP objective value and the global lower bound \\
		real    & relative position of the current node's LP objective value to the global upper/lower bounds \\
		real    & gap between the current node's LP objective value and the root node's upper bound \\
		real    & gap between the current upper bound and the root node's upper bound \\
		\bottomrule
	\end{tabular}
	\label{tab:global}
\end{table}%

\subsection {Graph Pointer Network Model}

\label{sec:model}

In this section, we propose a graph pointer network (GPN) similar to \cite{yang2022graph, ma2019combinatorial} that combines the graph neural network and the pointer mechanism to model the branching policy, which can map from the solver state to the branching decisions effectively.

From the features extracted in the previous section, it can be seen that the solver state has a bipartite graph structure, that is, the left nodes (variables) and the right nodes (constraints) are connected by edges, as shown in \fref{fig:bb2}. Graph neural network can effectively process the information of graph structure, and has been successfully applied to various machine learning tasks with graph structure input, such as social networks and citation networks. Therefore, we encode the graph structure of the solver state by a graph neural network model.

In addition, we take the global and historical features as a \emph{query}, and compute the attention value, which is then normalized as a softmax probability distribution, as a \emph{pointer} to the input sequence. In this way, the variable with the largest probability is selected as the branching variable.

The proposed graph pointer neural network model is composed of two parts: 1) the graph neural network calculates the feature vector for each variable based on variable features, constraint features and edge features; 2) the pointer mechanism outputs the variable selection probabilities by computing the attention values according to variables' feature vectors and the \emph{query} which is constructed by the global and historical features. The detailed process of modeling the branching policy is as follows.

(1) Initial embedding calculation

Variable features, constraint features, edge features, and global features have different dimensions. For example, the variable feature is 13-dimensional, and the global feature is 9-dimensional. Therefore, we first compute the $d_h$-dimensional embedding of the variable features $\mathbf{x}_{v}$, constraint features $\mathbf{x}_{c}$, edge features $\mathbf{x}_{e}$ and global features $\mathbf{x}_{g}$:
\begin{equation}
	\begin{array}{l}
		\mathbf{x}_{v} \leftarrow \operatorname{EMBEDDING}\left(\mathbf{x}_{v}\right) \\
		\mathbf{x}_{c} \leftarrow \operatorname{EMBEDDING}\left(\mathbf{x}_{c}\right) \\
		\mathbf{x}_{e} \leftarrow \operatorname{EMBEDDING}\left(\mathbf{x}_{e}\right) \\
		\mathbf{x}_{g} \leftarrow \operatorname{EMBEDDING}\left(\mathbf{x}_{g}\right)
	\end{array}
\end{equation}

where $\operatorname{EMBEDDING(\cdot)}$ is a two-layer fully connected neural network. The hidden dimension is $d_h$ and the activation function between layers is LeakyRELU:
\begin{equation}
	\text { LeakyRELU }(x)=\left\{\begin{array}{ll}
		x, & \text { if } x \geq 0 \\
		10^{-2} \times x, & \text { otherwise }
	\end{array}\right.
\end{equation}

(2) {Graph Neural Network}

Next, we compute the final variable features by a graph convolution neural network similar to \cite{gasse2019exact}:
\begin{equation}
	\begin{array}{l}
		\mathbf{x}_{c}^{i} \gets \mathbf{f}_\mathcal{C}\Big(\mathbf{x}_{c}^{i}, \sum_{j}^{(i, j) \in {E}}\mathbf{g}_\mathcal{C}\left(\mathbf{x}_{c}^{i}, \mathbf{x}_{v}^{j}, \mathbf{x}_{e}^{i,j}\right)\Big) \\
		\mathbf{x}_{v}^{j} \gets \mathbf{f}_\mathcal{V}\Big(\mathbf{x}_{v}^{j}, \sum_{j}^{(i, j) \in {E}}\mathbf{g}_\mathcal{V}\left(\mathbf{x}_{v}^{j}, \mathbf{x}_{c}^{i}, \mathbf{x}_{e}^{i,j}\right)\Big)
	\end{array}
\label{eq:graph}
\end{equation}

Function $\mathbf{g}(\cdot)$ is defined as:
\begin{equation}
	g\left(\mathbf{x}_{c}^{i}, \mathbf{x}_{v}^{j}, \mathbf{x}_{e}^{i,j}\right) = \operatorname{MLP}\left(\mathbf{x}_{c}^{i} + \mathbf{x}_{v}^{j} + \mathbf{x}_{e}^{i,j}\right)
\end{equation}

where $\operatorname{MLP}$ is a two-layer fully connected neural network with LeakyRELU activation function. Function $\mathbf{f}(\cdot)$ is also a two-layer fully connected neural network with LeakyRELU activation function. As demonstrated in \eref{eq:graph}, the graph embedding is computed by two successive convolution passes, one from variables to constraints and the next one from constraints to variables. The first convolution step computes the features $\mathbf{x}_{c}^{i}$ of constraint $i$ according to features $\mathbf{x}_{v}^{j}$ of its connected variables $j$, features of the edge $\mathbf{x}_{e}^{i,j}$ and its own features. The second step computes the embedding $\mathbf{x}_{v}^{j}$ of variable $j$ according to the above obtained features $\mathbf{x}_{c}^{i}$ of its connected constraints $i$, features of the edge $\mathbf{x}_{e}^{i,j}$ and its own features. Through the graph convolution process, the final variable features aggregate the original variable features, constraint features and coefficient features of the problem, so as to effectively contain the graph information of the MILP state.

(3) Historical feature calculation

At branching step $t$, the first part of the historical features is the past branching decisions $\mathcal{C}_1 = a_1 \cdots a_{t-1}$ at steps $1 \cdots t-1$. We compute this part of $d_h$-dimensional historical features as:
\begin{equation}
	\mathbf{x}_{h1}^t = \operatorname{MLP}\Big(\frac{1}{t-1} \sum_{i=1}^{t-1}\mathbf{x}_{v}^{a_{i}}\Big)
\end{equation}

where $\operatorname{MLP}$ is a single-layer fully connected neural network layer, and $a_i$ is the variable selected by the solver at step $i$.

The second part of the historical feature is the variable set $\mathcal{C}_2$ whose value changes during the process of generating the current node. The same operation is performed on $\mathcal{C}_2$ to obtain the $d_h$-dimensional vector $\mathbf{x}_{h2}^t$. In addition, $\mathbf{x}_{h2}^t$ and $\mathbf{x}_{h1}^t$ are zero vectors if $t==0$.

(4) {Pointer Mechanism}

We compute the attention value as a pointer to the candidate variables. The attention value is computed by a compatibility function of the \emph{query} with the \emph{key}.
The \emph{query}, which is composed of global features and historical features, represents the current state of the solver. The \emph{key} represents the feature of each candidate variable. In specific, the \emph{query} vector is calculated as the weighted average of global and historical features:
\begin{equation}
	\mathbf{q_t} = w_1*\mathbf{x}_{g}^t + w_2* \mathbf{x}_{h1}^t + w_3 *\mathbf{x}_{h2}^t
\end{equation}

where $w_1, w_2, w_3$ are weight values to be optimized while training. Moreover, the \emph{key} of variable $i$ is defined as $k_i = W_k \mathbf{x}_v^i, i \in \mathcal{C}$, which is the linear projection of the variable features. Denote the query at branching step $t$ as $\mathbf{q_t}$ and the keys of candidate variables as $\mathbf{k_i}, i \in \mathcal{C}$, one has:
\begin{equation}
	\begin{aligned}
		u_{i}^{t} &=W_3  \left(W_{1} k_{i}+W_{2} q_{t}\right) & i \in(1, \ldots, n) \\
		a_{i}^{t} &=\operatorname{softmax}\left(u_{i}^{t}\right) & i \in(1, \ldots, n)
	\end{aligned}
\end{equation}

where $u_{i}^{t}$ is the attention value computed by the compatibility function. Note that other compatibility function can also be applied to compute the attention, which can refer to \cite{vaswani2017attention} for more details. $\operatorname{softmax}$ is used to normalize the attention value to the probability distribution $a_{i}^{t}$, representing the probability of selecting variable $i$ at branching step $t$. In this case, we can choose the variable with the highest probability $a_{i}^{t}$ as the branching variable.

In addition, it is necessary to normalize the variable features, constraint features, edge features, and global features due to their different data range. To this end, we apply the prenorm layer as introduced in \cite{gasse2019exact} to normalize the variable, constraint, and edge features. We also add a prenorm layer of global features accordingly, so that the neural network model can deal with problem instance with global features of different scales.

\subsection {Branch and Bound algorithm based on GPN}

We use the GPN model to select the branching variable in $\mathrm{B}\&\mathrm{B}$. The GPN-based $\mathrm{B}\&\mathrm{B}$ algorithm is illustrated in algorithm \ref{alg:bb}.

\begin{algorithm}
	\small
	\caption{Branch and Bound algorithm based on GPN}
	\label{alg:bb}
	\begin{algorithmic}[2]
		\REQUIRE {Root Node $R$, representing the LP relaxation of the original MILP}
		\ENSURE {Optimal solution $S^*$}
		\STATE $R.lowerBound \leftarrow -\infty$ \quad //Initialize the lower bound of $R$
		\STATE $Queue \leftarrow \{ R \}$ \quad //Store the unexplored node into the Queue
		\STATE $UpperBound \leftarrow \infty$ \quad //Initialize the global upper bound
		\STATE $S^* \leftarrow null$ \quad
		\WHILE{$Queue$ is not empty}
		\STATE $N \leftarrow Queue.get()$ \quad // Dequeue the node 		
		\IF{$N.lowerBound \geq UpperBound$}
		\STATE // If node $N$'s parent node's lower bound is greater than the global upper bound, prune this node
		\STATE continue
		\ENDIF
		\STATE $S_{r} \leftarrow \operatorname{solve}(N)$
		\IF{$S_r$ is not feasible}
		\STATE // prune this node
		\STATE continue
		\ENDIF
		\STATE $O_{r} \leftarrow S_r.objectiveValue$
		\IF{$O_{r} > UpperBound$}
		\STATE // If node $N$'s lower bound is greater than the global upper bound, prune this node
		\STATE continue
		\ENDIF
		\IF{$S_r$ is feasible}
		\STATE $UpperBound \leftarrow O_{r}$
		\STATE $S^* \leftarrow S_{r}$
		\STATE // Update the global upper bound and $S^*$
		\STATE continue
		\ENDIF
		\STATE Extract features of the solver state, $state=\left( V,C,E,G,H\right)$
		\STATE $V \leftarrow \operatorname{GPN}(S_r, state)$ \quad //Select varibale $V$ by the GPN model
		\STATE $a \leftarrow \operatorname{floor}(V.value)$
		\STATE $L \leftarrow \operatorname{addConstraint}(N,V \leq a)$
		\STATE $R \leftarrow \operatorname{addConstraint}(N,V \geq a)$
		\STATE // Branch on $V$ and obtain the two LP sub-porblems
		\STATE $L.lowerBound \leftarrow O_r$, $R.lowerBound \leftarrow O_r$
		
		\STATE $Queue.add(L)$, $Queue.add(R)$
		
		\ENDWHILE
		\RETURN $S^*$
	\end{algorithmic}
\end{algorithm}

First, the LP relaxation of the original MILP problem is set as the root node. The queue data structure is maintained to store the sub-problem nodes to be solved. Each node defines an initial lower bound $l$, which represents the lower bound of its parent node. After the global upper bound is updated, if $l$ is greater than the global upper bound, then the node will be pruned. When the node is taken out of the queue, its lower bound is compared with the global upper bound, and the node is pruned if the lower bound is greater than the global upper bound. The global upper bound is initialized to $\infty$, and is updated every time a better feasible solution is obtained. We extract variable, constraint, edge, global and historical features of candidate variables, which are subsequently input to the GPN model. The model output the probability distribution of the candidate variables. We can select the one with the highest probability as the variable to branch on. And  two sub-problems are generated accordingly. This process loops until the queue is empty, i.e., all leaves of the search tree are explored.

\section {Training method}
\label{sec:policy-learning}
We use imitation learning to train the proposed model. The objective is to imitate the strong branching rule.
Imitation learning \cite{hussein2017imitation} can solve various multi-step decision-making problems. In comparison with unsupervised reinforcement learning methods, imitation learning can improve the training efficiency with the help of expert experiences. Imitation learning requires labeled training data provided by human experts $\left\{\tau_{1}, \tau_{2}, \ldots, \tau_{m}\right\}$, where $\tau_{i}=<s_{1}^{i}, a_{1}^{i}, s_{2}^{i}, a_{2}^{i}, \ldots>$. $s_{1}^{i}, a_{1}^{i}$ represents the ``state-action" pairs in a Markov decision process generated by solving an instance using the SB-based $\mathrm{B}\&\mathrm{B}$. Therefore, the labeled training set can be constructed as $\mathcal{D}=\left\{\left(s_{1}, a_{1}\right),\left(s_{2}, a_{2}\right),\left(s_{3}, a_{3}\right), \ldots\right\}$. Denote $a_{i}$ as the label, the variable selection problem can be converted into a classification problem. The objective is to minimize the difference between the expert actions and the predicted actions.

In specific, we conduct the SB-based $\mathrm{B}\&\mathrm{B}$ on randomly generated combinatorial optimization instances. The ``state-action" pairs are recorded to form a training set $\mathcal{D}=\{(\mathbf{s}_i, \mathbf{a}^\star_i)\}_{i=1}^N$. Denote the expert actions as $\mathbf{a}^\star$ and the predicted actions as $\pi(s)$, we optimize the model parameters $\theta$ by minimizing:
\begin{equation}
	\label{eq:loss}
	\mathcal{L}(\theta) = \frac{1}{N}\sum_{(\mathbf{s}, \mathbf{a}^*)\in \mathcal{D}} \operatorname{loss} \left( \pi_\theta(\mathbf{s}), \mathbf{a}^\star \right).
\end{equation}
where $\operatorname{loss}(*)$ is a function that defines the difference between the true value and the predicted value. For classification problems, there are a number of $\operatorname{loss}(*)$ functions such as the accuracy and cross entropy.

However, in $\mathrm{B}\&\mathrm{B}$, SB scores of different variables might be the same or pretty close. It is equivalent to select these variables. In this case, we record the SB scores instead of the variable indices to construct the training set $\mathcal{D}=\{(\mathbf{s}_i, {score}^\star_i)\}_{i=1}^N$. The aim is to imitate the distribution of the SB scores instead of the branching actions. To this end, we use the Kullback-Leibler (KL) divergence as a measure of the difference between the SB score distribution and the predicted probability distribution. By minimizing the KL divergence, the model can can work better for the above situation where multiple variables own the same or similar SB scores.

Denote $P$ as the true distribution of the data and $Q$ as the predicted distribution of the model to fit $P$, KL divergence is defined as:
\begin{equation}
	D_{\mathrm{KL}}(P \| Q)=\sum_{x \in \mathcal{X}} P(x) \log \left(\frac{P(x)}{Q(x)}\right)
	\label{eq:kl}
\end{equation}

Therefore, we optimize the model parameters $\theta$ by minimizing:
\begin{equation}
	\mathcal{L}(\theta) = D_{\mathrm{KL}}({score}^\star \| \pi_\theta(\mathbf{s})) = \sum_{(\mathbf{s},  {score}^\star)\in \mathcal{D}} {score}^\star \log \left(\frac{{score}^\star}{\pi_\theta(\mathbf{s})}\right)
\end{equation}

In addition, we only care about the variables with high SB scores. The probability distribution of other variables has no effect on the branching variable selection. Thereby, we emphasize the similarity loss of variables with high SB scores in the training phase. In specific, we sort the variables according to their probabilities output by the model. We should pay more attention to the first few variables. To this end, the KL divergence of the top-k variables is added to the loss item. We first sort the probabilities $\pi_\theta(\mathbf{s})$ output by the model, and select the first $k$ variables $\mathcal{I}_k$. The KL divergence value of variables $\mathcal{I}_k$ is computed by \eref{eq:kl} as $D_{\mathrm{KL}}({score}_{\mathcal{I}_k}^\star \| \pi_\theta(\mathbf{s})_{\mathcal{I}_k})$. And the loss for training the model is defined as:

\begin{equation}
	\mathcal{L}(\theta) = D_{\mathrm{KL}}({score}^\star \| \pi_\theta(\mathbf{s})) + D_{\mathrm{KL}}({score}_{\mathcal{I}_k}^\star \| \pi_\theta(\mathbf{s})_{\mathcal{I}_k})
\end{equation}

The first term of the loss can make the overall predicted distribution similar to the distribution of the SB scores, while the second term makes the model pay more attention to the variables of large probabilities and weaken the distribution of irrelevant variables for selecting the branching variables. This can alleviate the situation where a large amount of training time is cost to fit the distribution of irrelevant variables.

\section {Experimental Results and Discussion}

\subsection {Experiment Settings}
\label{sec:experiments}

\subsubsection {Comparison Algorithm}
We compare the proposed approach against the following approaches:

(1) First, we compare the proposed approach against the classic $\mathrm{B}\&\mathrm{B}$ algorithm. The branching rule of reliability branching (RB), strong branching (SB) and pseudocost branching (PB) are compared respectively. They are all implemented in the well-known SCIP solver. The cutting plane is only allowed at the root node. Other heuristics are disabled during the branching process for fair comparison. Our method is also implemented in the SCIP solver, and uses the same set of parameters as the competitor methods.

(2) Next, the proposed approach is compared with the state-of-the-art machine learning-based $\mathrm{B}\&\mathrm{B}$ algorithms: branching method based on ExtraTrees \cite{geurts2006extremely} model \cite{alvarez2017machine} ({TREES}); branching method \cite{Khalil2016} ({ SVMRANK}) and \cite{hansknecht2018cuts} ({LMART}) based on SVMrank \cite{joachims2002optimizing} and LambdaMART \cite{burges2010ranknet} model; branching method based on graph neural network \cite{gasse2019exact}({ GNN}).

\subsubsection {Test Problems}
\label{sec:setup}

Effectiveness of the proposed method is evaluated on the following three benchmark combinatorial optimization problems.

(1) Set covering problem \cite{balas1980set}

The set covering instances contain 1,000 columns. The model is trained on instances with 500 rows, and is evaluated on instances with 500 and 1,000 rows, respectively.

(2) Capacitated facility location problem \cite{cornuejols1991comparison}

The instances are generated with 100 facilities. The model is trained on instances with 100 customers, and is evaluated on instances with 100 and 200 customers, respectively.

(3) Maximum independent set problem \cite{bergman2016decision}

The instances are generated following the process in \cite{bergman2016decision}. The model is trained on instances of 500 nodes, and is evaluated on instances with 500 and 1000 nodes, respectively.

\subsubsection{Experimental parameter settings}
All compared algorithms are implemented by Python on the SCIP solver. SCIP uses its default parameters. The hidden dimensions of the models are set to $d_h=64$. The Adam optimizer is used for training with learning rate of 0.001. We set $k=10$ for the top-k imitation learning. The learning rate decreases 80\% if the loss does not decrease for 10 epochs. The training is terminated if the loss does not decrease for 20 epochs.
\subsubsection {Training and evaluation}

(1){Training data generation}

The SCIP solver with default settings is used to collect training samples offline. We generate random instances and solve them using the SCIP. During the collecting procedure, the branching rule of RB is adopted with a probability of 95\%, and the branching rule of SB is adopted with a probability of 5\%. Only the samples generated by SB are collected. The data of variable, constraint, edge, global and historical features, candidate variable sets, and SB scores of the variables is collected.

Instances are randomly generated and solved until 140,000 samples are collected. 100,000 samples are used as the training set, 2,000 samples are used as the validation set, and 2,000 samples are used as the test set.

(2) Evaluation method

We first evaluate the capability of the GPN method in imitating the SB rule. Since multiple variables may have the same or similar SB scores, the following indices are used to evaluate the model accuracy \cite{gasse2019exact}: 1) the percentage of times the output of the model is exactly the variable with the highest SB score (acc@1); 2) the percentage of times the output of the model is one of the five variables with the highest SB scores (acc@5); 3) the percentage of times the output of the model is one of the ten variables with the highest SB scores (acc@10). Moreover, we evaluate the total solving time of the GPN-based $\mathrm{B}\&\mathrm{B}$ in comparison with benchmark methods.

\subsection {Results}

\begin{figure}[htbp]
	\centering
	\subfloat[Loss]{\includegraphics[width=2.2in]{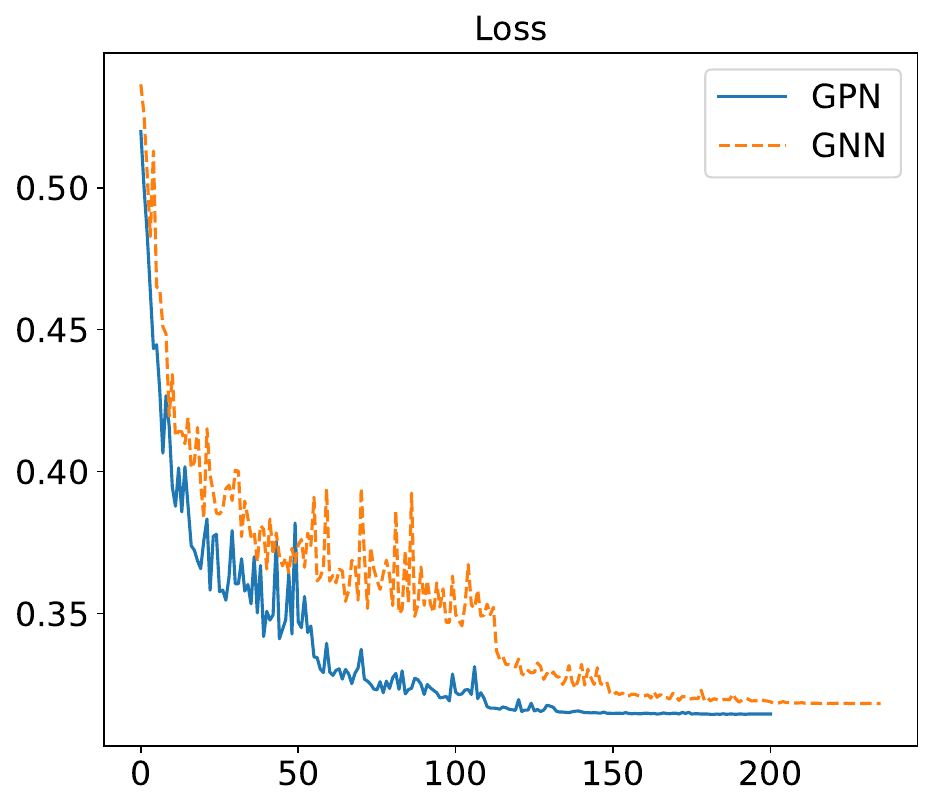}
	}
	\hfil
	\subfloat[Model accuracy]{\includegraphics[width=2.2in]{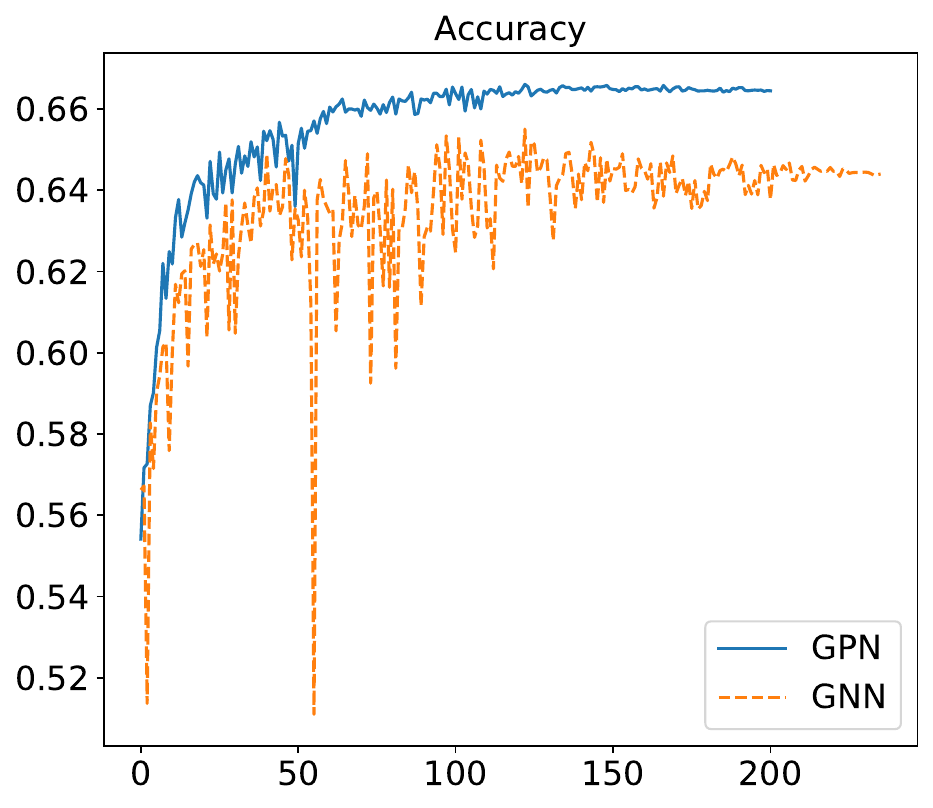}
	}
	\caption{Training performances of the models on set covering.}
	\label{fig:set}
\end{figure}

\begin{figure}[htbp]
	\centering
	\subfloat[Loss]{\includegraphics[width=2.2in]{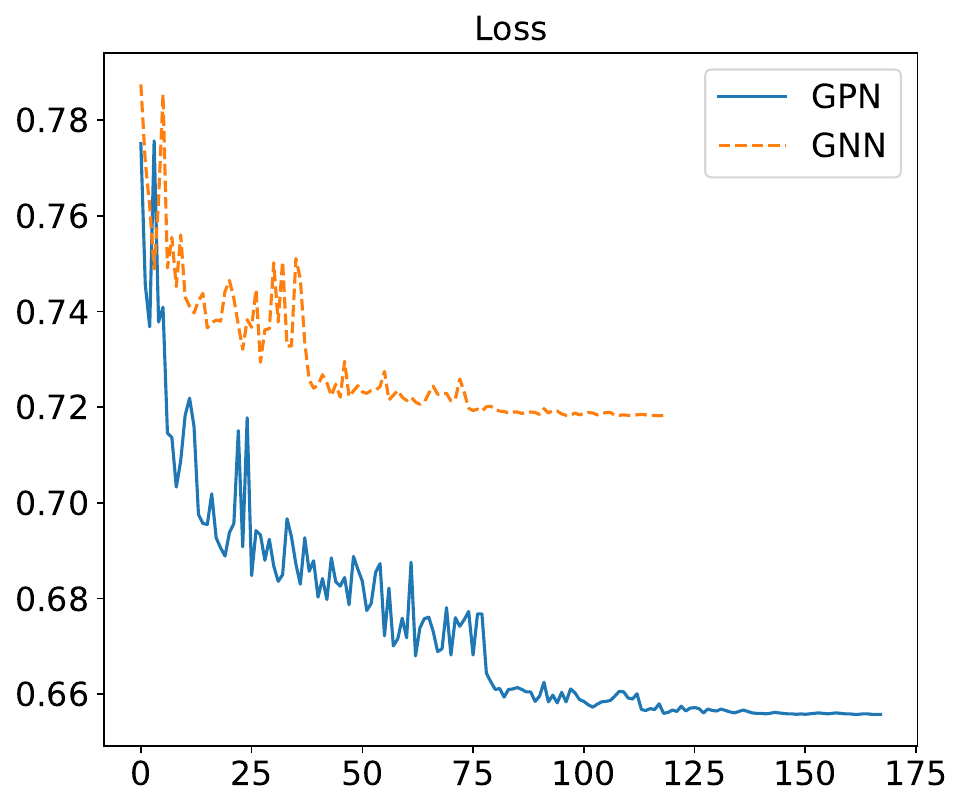}
	}
	\hfil
	\subfloat[Model accuracy]{\includegraphics[width=2.2in]{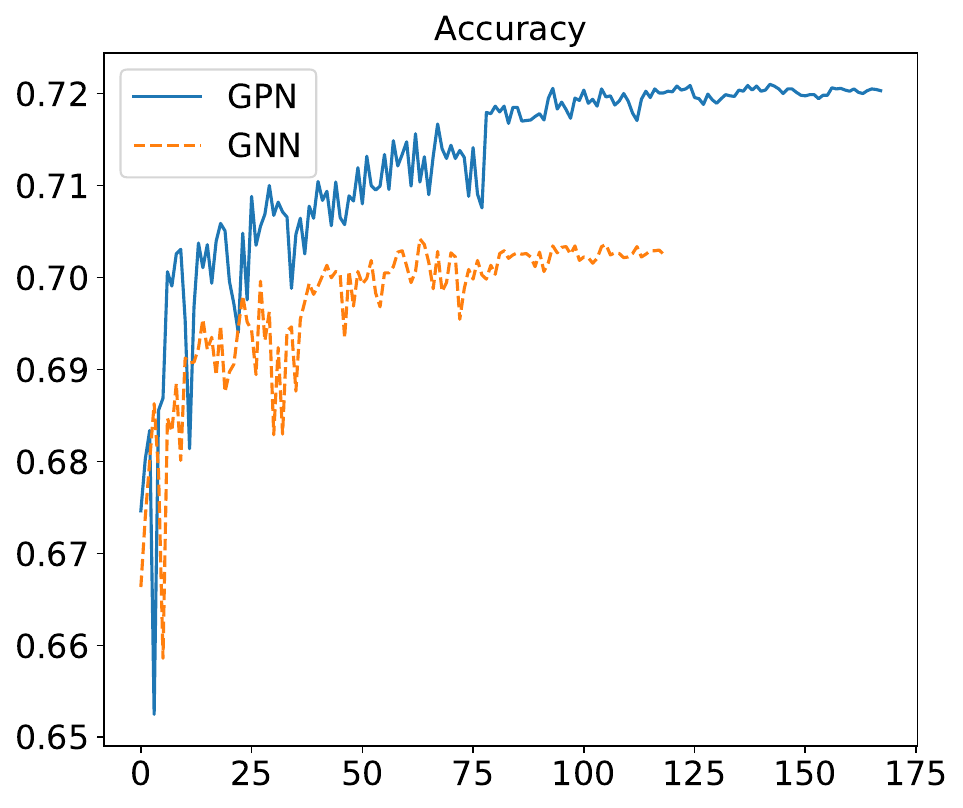}
	}
	\caption{Training performances of the models on capacitated facility location.}
	\label{fig:loc}
\end{figure}

\begin{figure}[htbp]
	\centering
	\subfloat[Loss]{\includegraphics[width=2.2in]{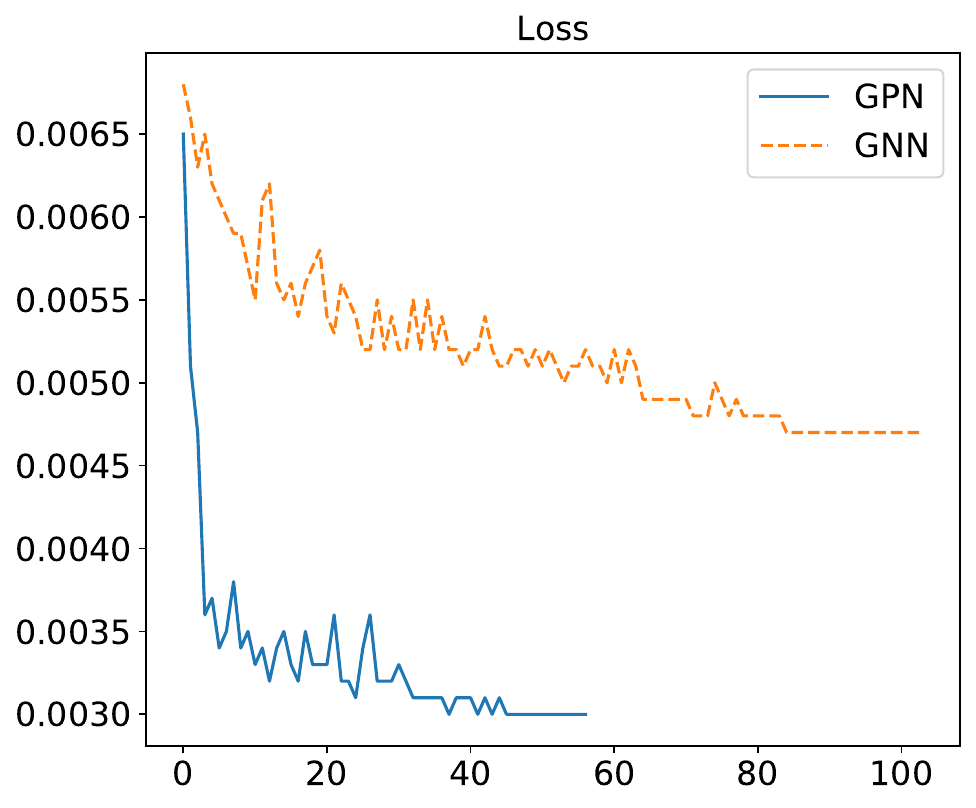}
	}
	\hfil
	\subfloat[Model accuracy]{\includegraphics[width=2.2in]{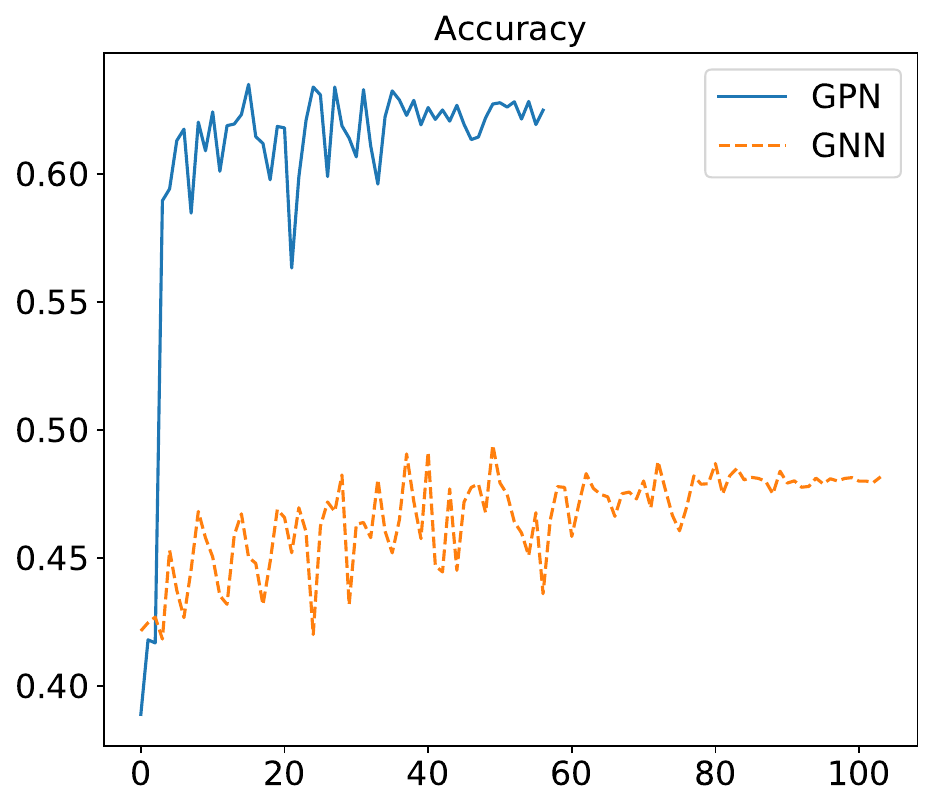}
	}
	\caption{Training performances of the models on maximum independent set.}
	\label{fig:ind}
\end{figure}

\fref{fig:set}, \fref{fig:loc} and \fref{fig:ind} present the training performances of the proposed GPN model and the classic GNN model on three test problems. The convergence of the loss and model accuracy on the validation set are compared.

GNN is currently a benchmark model for imitating the branching rule. Results show that the proposed GPN model outperforms the traditional GNN model in terms of both convergence speed and final convergence performance while training. Moreover, GPN comfortably outperforms GNN in terms of the model accuracy for all of the three problems on validation set. The advantages of the GPN are more obvious on the location problem and the maximum independent set problem, where the GPN can converge to 0.66 and 0.003 while the GNN can only converge to 0.72 and 0.0047. This validates the effectiveness of the proposed GPN model. The attention-based pointer mechanism proposed in the GPN can effectively understand the graph and global characteristics of the problem, thus making more accurate decisions.

\tref{tab:set}, \tref{tab:location}, \tref{tab:ind} present the model accuracy of GPN, TREES, SVMRANK, LMART, and GNN methods on test set. Results of TREES, SVMRANK and LMART are from \cite{gasse2019exact}. Results of acc@1, acc@5 and acc@10 are listed respectively.

\begin{table}[htbp]
	\centering
	\caption{Results of model accuracy on set covering.}
	\begin{tabular}{lrrr}
		\toprule
		& \multicolumn{1}{l}{acc@1} & \multicolumn{1}{l}{acc@5} & \multicolumn{1}{l}{acc@10} \\
		\midrule
		TREES & 51.8  & 80.5  & 91.4 \\
		SVMRANK & 57.6  & 84.7  & 94 \\
		LMART & 57.4  & 84.5  & 93.8 \\
		GNN   & 65.5  & 92.4  & \textbf{98.2} \\
		GPN   & \textbf{66.5} & \textbf{92.7} & \textbf{98.2} \\
		\bottomrule
	\end{tabular}
	\label{tab:set}
\end{table}%

\begin{table}[htbp]
	\centering
	\caption{Results of model accuracy on capacitated facility location.}
	\begin{tabular}{lrrr}
		\toprule
		& \multicolumn{1}{l}{acc@1} & \multicolumn{1}{l}{acc@5} & \multicolumn{1}{l}{acc@10} \\
		\midrule
		TREES & 63    & 97.3  & \textbf{99.9} \\
		SVMRANK & 67.8  & 98.1  & \textbf{99.9} \\
		LMART & 68    & 98    & \textbf{99.9} \\
		GNN   & 71.2  & 98.6  & \textbf{99.9} \\
		GPN   & \textbf{72.2} & \textbf{98.7} & \textbf{99.9} \\
		\bottomrule
	\end{tabular}
	\label{tab:location}
\end{table}%

\begin{table}[htbp]
	\centering
	\caption{Results of model accuracy on maximum independent set.}
	\begin{tabular}{lrrr}
		\toprule
		& \multicolumn{1}{l}{acc@1} & \multicolumn{1}{l}{acc@5} & \multicolumn{1}{l}{acc@10} \\
		\midrule
		TREES & 30.9  & 47.4  & 54.6 \\
		SVMRANK & 48    & 69.3  & 78.1 \\
		LMART & 48.9  & 68.9  & 77 \\
		GNN   & 56.5  & 80.8  & 89 \\
		GPN   & \textbf{63.2} & \textbf{86.9} & \textbf{92.6} \\
		\bottomrule
	\end{tabular}
	\label{tab:ind}
\end{table}%

It is observed that the GPN method has the highest accuracy among the compared approaches on all the three test problems. Its advantage over traditional machine learning methods is more obvious on the maximum independent set problem. We can see a significant effectiveness of the GPN model.

In addition, we evaluate the running time of the approaches, since the aim of the branching models is to reduce the overall solving time of $\mathrm{B}\&\mathrm{B}$. The solving time is determined by the size of the search tree, that is, the number of explored nodes. It is also determined by the time consumed by making the branch decisions. Therefore, a good branching model can reduce the size of the search tree while making fast branching decisions.

\tref{tab:s1}, \tref{tab:s3}, \tref{tab:s3} list the results of solving time and the number of explored nodes when using GPN and the compared approaches for solving the three test problems. Results are obtained by solving 100 randomly generated problems and taking the average.

\tref{tab:s1} shows that, in comparison with the PB and RB rule, the proposed GPN method achieves at least 40\% increase in the solution speed when solving the 500- and 1000-row set covering instances. In terms of the number of explored nodes, GPN outperforms all of the compared methods except for the SB rule on the set cover instances. SB can always get the smallest search tree. But its total solving time has no advantage due to its long computation time of making branching decisions. It is obvious that the GPN method outperforms all the compared machine learning methods in terms of the solving speed and the ability of reducing the search tree on the set covering instances.

It can be seen from \tref{tab:s2} that the GPN method shows greater advantages in solving the 100- and 200-customer capacitated facility location instances than the compared methods. In specific, GPN runs twice faster than the PB and RB method. Compared with the machine learning methods, GPN has the fastest solving speed and the fewest number of nodes.

On maximum independent set instances, GPN achieves nearly 10\% improvement in the solution speed and 20\% reduction in the number of nodes as seen in \tref{tab:s3}. The solving time is reduced nearly twice when using the GPN compared with the PB and RB methods.

Note that, the test instances are generated randomly, and are different from the training set. Once the model is trained, it can generalize to unseen instances, and scale to larger instances. Although the RB heuristic is carefully handcrafted by experts, it is still defeated by the proposed GPN method, which can learn the heuristics from data. Experiments validate the novelty and efficiency of the GPN method.

\begin{table}[htbp]
	\centering
	\caption{Results of running time on set covering.}
	\begin{tabular}{lrr|rr}
		\toprule
		& \multicolumn{2}{c|}{500 rows} & \multicolumn{2}{c}{1000 rows} \\
		\midrule
		Methods    & \multicolumn{1}{l}{Time} & \multicolumn{1}{l|}{Nodes} & \multicolumn{1}{l}{Time} & \multicolumn{1}{l}{Nodes} \\
		\midrule
		SB    & 7.02  & \textbf{12.5} & 173.9 & \textbf{227.5} \\
		PB    & 2.88  & 98.6  & 19.8  & 2211.2 \\
		RB   & 3.73  & 18.8  & 22.1  & 1192.5 \\
		GNN   & 2.07  & 43.9  & 14.2  & 900.6 \\
		GPN   & \textbf{2.04} & 42.3  & \textbf{13.9} & 891.2 \\
		\bottomrule
	\end{tabular}
	\label{tab:s1}
\end{table}%

\begin{table}[htbp]
	\centering
	\caption{Results of running time on capacitated facility location.}
	\begin{tabular}{lrr|rr}
		\toprule
		& \multicolumn{2}{c|}{100 customers} & \multicolumn{2}{c}{200 customers} \\
		\midrule
		Methods    & \multicolumn{1}{l}{Time} & \multicolumn{1}{l|}{Nodes} & \multicolumn{1}{l}{Time} & \multicolumn{1}{l}{Nodes} \\
		\midrule
		SB    & 157.4 & \textbf{116.5} & 1163.3 & \textbf{158.7} \\
		PB    & 82.8  & 541.9 & 510.7 & 614.2 \\
		RB   & 96.7  & 264.7 & 598.9 & 303.5 \\
		GNN   & 37.4  & 467.4 &   145.6    &  529.6\\
		GPN   & \textbf{35.2}  & 428.9 &   \textbf{140.3}    & 516.2 \\
		\bottomrule
	\end{tabular}
	\label{tab:s2}
\end{table}%

\begin{table}[htbp]
	\centering
	\caption{Results of running time on maximum independent set.}
	\begin{tabular}{lrr|rr}
		\toprule
		& \multicolumn{2}{c|}{500 nodes} & \multicolumn{2}{c}{1000 nodes} \\
		\midrule
		Methods    & \multicolumn{1}{l}{Time} & \multicolumn{1}{l|}{Nodes} & \multicolumn{1}{l}{Time} & \multicolumn{1}{l}{Nodes} \\
		\midrule
		SB    & 87.1  & \textbf{35.41} & 2844.4 & \textbf{164.5} \\
		PB    & 14.6  & 1937.8 & 2002.9 & 17213 \\
		RB   & 11.4  & 92.7  & 210.2 & 6717 \\
		GNN   & 5.01  & 61.7  &  222.5    &  14862 \\
		GPN   & \textbf{4.63} & 45.3  &   \textbf{198.6}    & 12587 \\
		\bottomrule
	\end{tabular}
	\label{tab:s3}
\end{table}%

The goal of $\mathrm{B}\&\mathrm{B}$ is to solve the combinatorial optimization problem as fast as possible, so the branch strategy must be a trade-off between the quality of the decision and the time spent on each decision. An extreme example is the SB branch rule, by calculating the SB score for variable selection, the final solution can be obtained with a small number of searches, but each decision step is very time-consuming, so that the overall running time is very long. Therefore, the method proposed in this study can achieve better decision quality and decision time. The trade-off of the SB score is slightly worse than the SB score, but it requires less calculation time, thus improving the overall solution speed.

\section {Conclusion}

This paper modeled the variable selection strategy in $\mathrm{B}\&\mathrm{B}$ with a deep neural network model. In addition to graph features, we further designed the global and historical features to represent the solver state. The model is comprised of the graph neural network and the pointer mechanism. Graph neural network is used to encode the graph features as the queries for the pointer. The global and historical features are processed as the key. The attention value is computed by the queries and the key as a pointer to the input sequence. We demonstrate on benchmark problems that, our approach can improve the overall $\mathrm{B}\&\mathrm{B}$ performance over traditional expert-deigned branching rules. Our approach can also outperform the state-of-the-art machine-learning-based $\mathrm{B}\&\mathrm{B}$ methods.

In future work, more combinatorial problems should be investigated via the proposed GPN model. Reinforcement learning methods can also be studied to improve the models trained by imitation learning.


%



\ifCLASSOPTIONcaptionsoff
  \newpage
\fi



\bibliographystyle{IEEEtran}
\bibliography{my}

\begin{IEEEbiography}[{\includegraphics[width=1in,height=1.25in,clip,keepaspectratio]{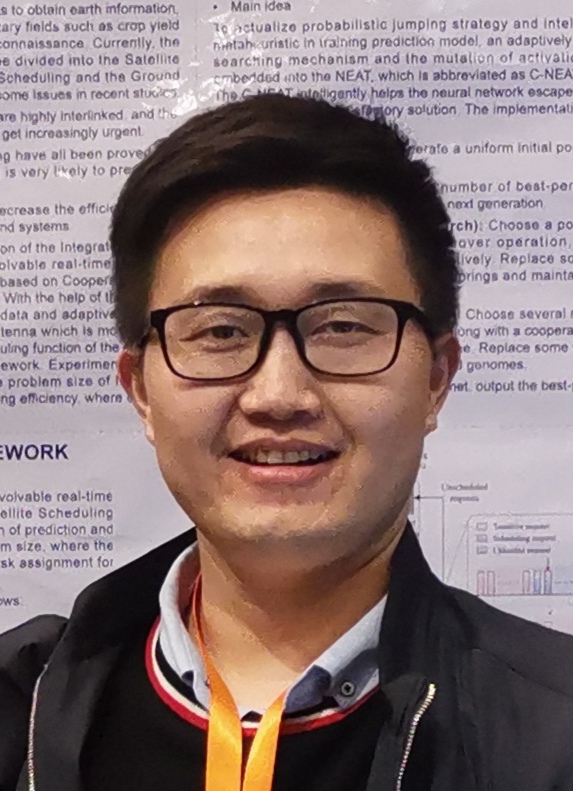}}]{Rui Wang} received his Bachelor degree from the National University of Defense Technology, P.R. China in 2008, and the Doctor degree from the University of Sheffield, U.K in 2013. Currently, he is an Associate professor with the National University of Defense Technology. His current research interest includes evolutionary computation, multi-objective optimization and the development of algorithms applicable in practice.
Dr. Wang received the Operational Research Society Ph.D. Prize at 2016, and the National Science Fund for Outstanding Young Scholars at 2021. He is also an Associate Editor of the Swarm and Evolutionary Computation, the IEEE Trans on Evolutionary Computation.
\end{IEEEbiography}

\begin{IEEEbiography}[{\includegraphics[width=1in,height=1.25in,clip,keepaspectratio]{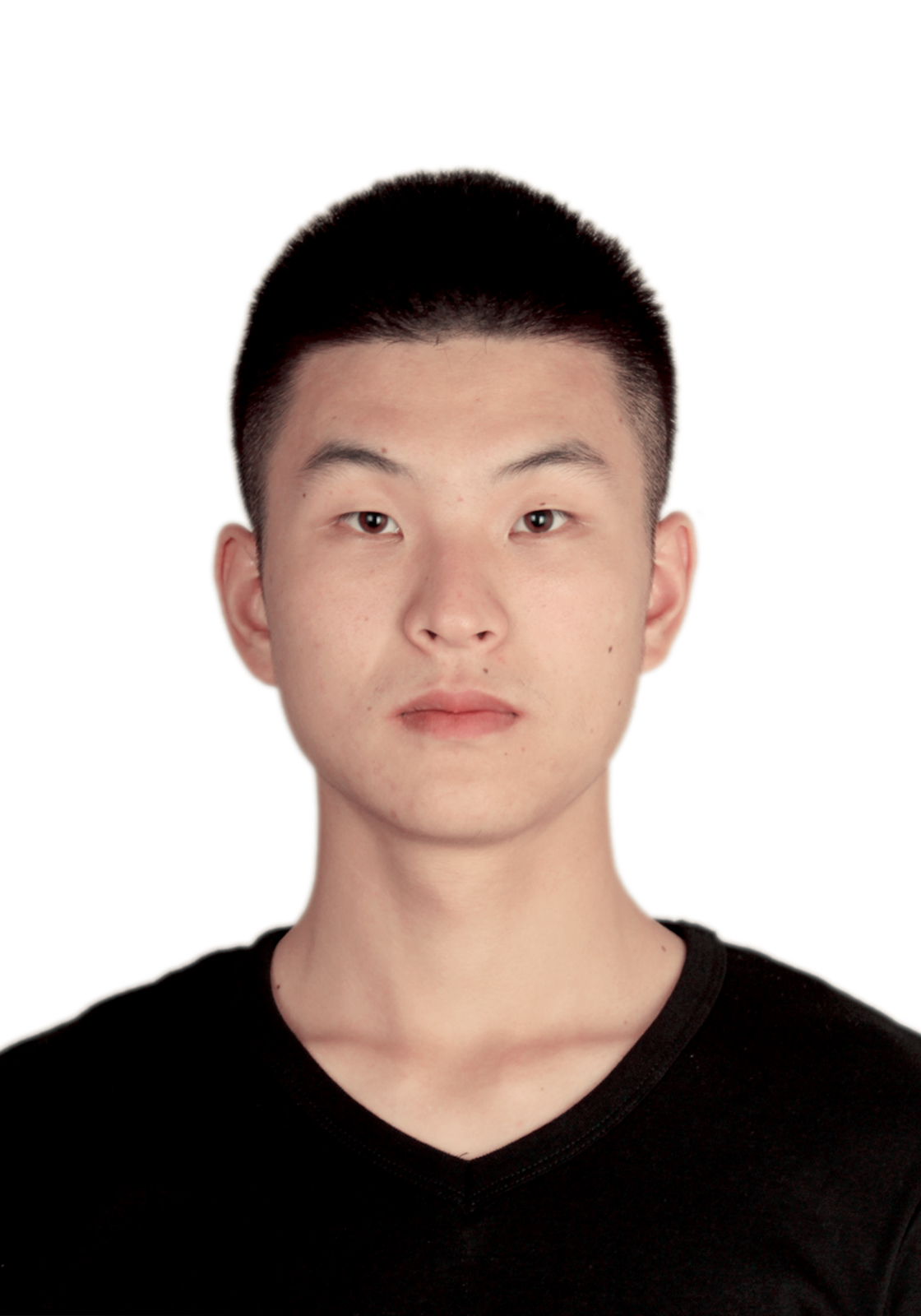}}]{Kaiwen Li} received the B.S., M.S. degrees from National University of Defense Technology (NUDT), Changsha, China, in 2016 and 2018.
	
He is a student with the College of Systems Engineering, NUDT. His research interests include prediction technique, multiobjective optimization, reinforcement learning, data mining, and optimization methods on Energy Internet.
\end{IEEEbiography}

\begin{IEEEbiography}[{\includegraphics[width=1in,height=1.25in,clip,keepaspectratio]{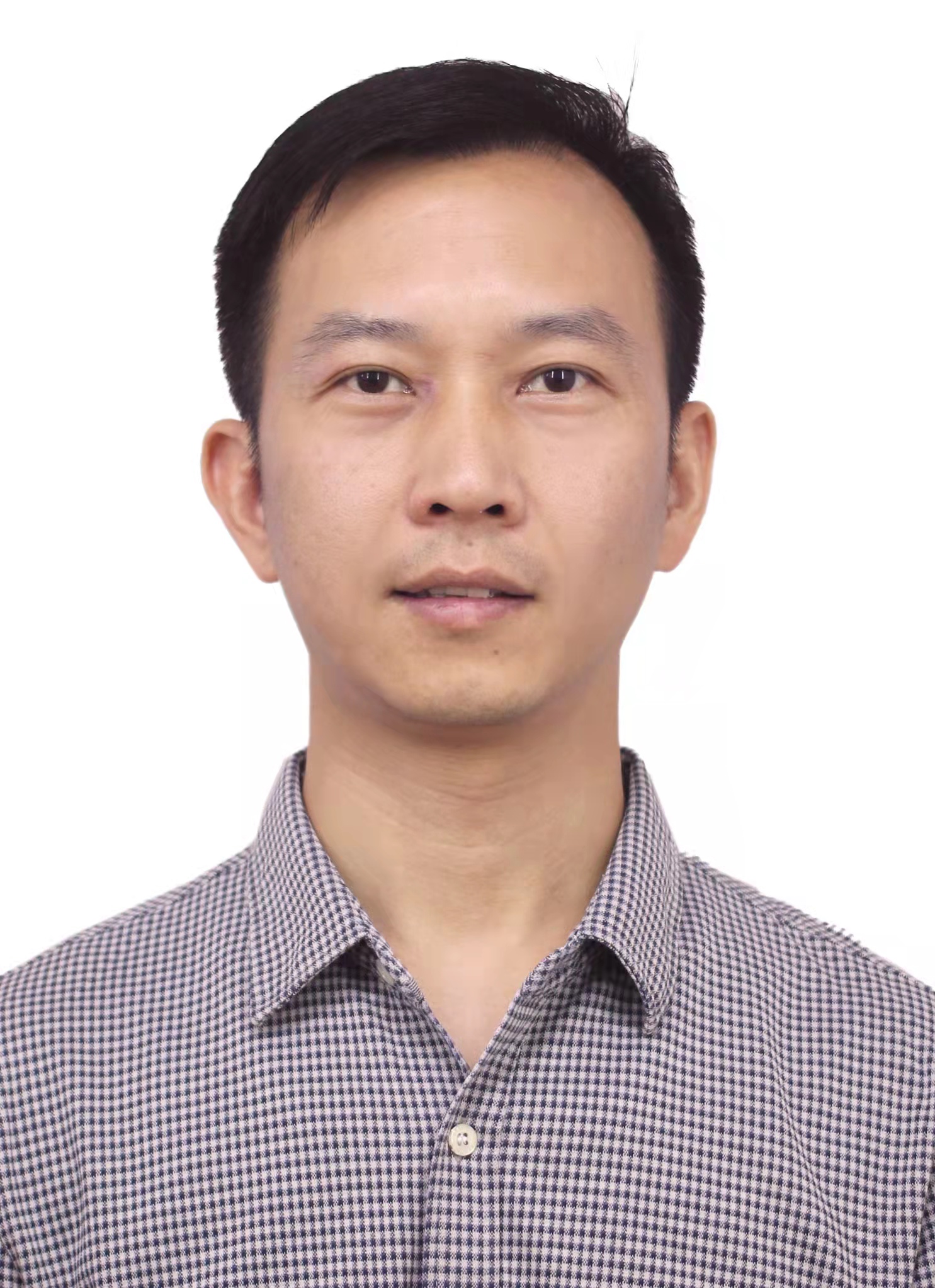}}]{Tao Zhang} received the B.S., M.S., Ph.D. degrees from National University of Defense Technology (NUDT), Changsha, China, in 1998, 2001, and 2004, respectively.
	
He is a Professor with the College of Systems Engineering, NUDT. His research interests include multicriteria decision making, optimal scheduling, data mining, and optimization methods on energy Internet network.
\end{IEEEbiography}

\begin{IEEEbiography}[{\includegraphics[width=1in,height=1.25in,clip,keepaspectratio]{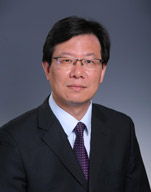}}]{Ling Wang} received his B.Sc. in automation and Ph.D. degree in control theory and control engineering from Tsinghua University, Beijing, China, in 1995 and 1999, respectively. Since 1999, he has been with the Department of Automation, Tsinghua University, where he became a Full Professor in 2008. His current research interests include intelligent optimization and production scheduling. He was the recipient of the National Natural Science Fund for Distinguished Young Scholars of China, the National Natural Science Award (second place) in 2014, the Science and Technology Award of Beijing City in 2008, and the Natural Science Award (first place in 2003, and second place in 2007) nominated by the Ministry of Education of China.
\end{IEEEbiography}

\begin{IEEEbiography}[{\includegraphics[width=1in,height=1.25in,clip,keepaspectratio]{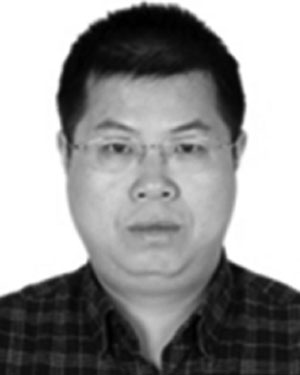}}]{Xiangke Liao} received the BS degree from Tsinghua University, China, in 1985, and the MS degree from the National University of Defense Technology (NUDT), China, in 1988, both in computer science. He is currently a professor with the College of Computer, NUDT. His research interests include high-performance computing systems, operating systems, and parallel and distributed
computing.He is the principle investigator and chief designer of Tianhe-2 supercomputer.
\end{IEEEbiography}




\end{document}